\documentclass{article}

\newif\ifcomments
\commentstrue

\usepackage[utf8]{inputenc} % allow utf-8 input

\usepackage{microtype}
\usepackage{graphicx}
\usepackage{subfigure}
\usepackage{times}
\usepackage{pdfcomment}
\usepackage{booktabs} % for professional tables
\usepackage{hyperref}
\usepackage[accepted]{icml2018}

\usepackage{color}
\usepackage[dvipsnames]{xcolor}
\usepackage{amsmath,amsthm,verbatim,amssymb,amsfonts,amscd,mathrsfs}
\usepackage[algo2e,ruled]{algorithm2e}
\usepackage{multirow}
\usepackage{makecell}  % thead, other vertical alignment options
\usepackage{hhline}  % double horizontal lines for tables

\usepackage{url}

\makeatletter
\DeclareRobustCommand*\textsubscript[1]{%
  \@textsubscript{\selectfont#1}}
\def\@textsubscript#1{%
  {\m@th\ensuremath{_{\mbox{\fontsize\sf@size\z@#1}}}}}
\makeatother

\ifcomments\newcommand{\comments}[1]{#1}\else\newcommand{\comments}[1]{}\fi
\definecolor{clrgp}{rgb}{.9,0,.9}

\usepackage{xcolor}
\definecolor{dark-blue}{rgb}{0.15,0.15,0.4}
\definecolor{medium-blue}{rgb}{0,0,0.5}
\hypersetup{
   colorlinks, linkcolor={dark-blue},
   citecolor={dark-blue}, urlcolor={black}
}

\newcommand{\reals}{\ensuremath{\mathbb{R}}}

\newcommand{\normaldist}[2]{\ensuremath{\mathcal{N} \left( #1, #2 \right)}}
\newcommand{\bigo}[1]{\mathcal O ( #1 )}

\newif\ifboldmatrix
\boldmatrixfalse
\ifboldmatrix\newcommand{\boldmatrix}[1]{\mathbf{#1}}\else\newcommand{\boldmatrix}[1]{#1}\fi

\newcommand{\ba}{\ensuremath{\mathbf{a}}}
\newcommand{\be}{\ensuremath{\mathbf{e}}}
\newcommand{\x}{\ensuremath{\mathbf{x}}}
\newcommand{\bb}{\ensuremath{\mathbf{b}}}
\newcommand{\bk}{\ensuremath{\mathbf{k}}}
\newcommand{\bv}{\ensuremath{\mathbf{v}}}
\newcommand{\bq}{\ensuremath{\mathbf{q}}}
\newcommand{\bw}{\ensuremath{\mathbf{w}}}

\newcommand{\bu}{\ensuremath{\mathbf{u}}}

\newcommand{\X}{\ensuremath{\boldmatrix{X}}}

\newcommand{\W}{\ensuremath{\boldmatrix{W}}}
\newcommand{\B}{\ensuremath{\boldmatrix{B}}}
\newcommand{\A}{\ensuremath{\boldmatrix{A}}}

\newcommand{\K}{\ensuremath{\boldmatrix{K}}}
\newcommand{\y}{\ensuremath{\mathbf{y}}}

\newcommand{\fn}{\ensuremath{\mathbf{f}}}

\newcommand{\dset}{\ensuremath{\mathcal D}}
\newcommand{\eye}{\ensuremath{I}}
\newcommand{\posterior}{\ensuremath{\! \mid \! \dset}}

\newcommand{\methodname}{LOVE}
\newcommand{\ourmethodcolor}{Magenta}

\newcommand{\blue}[1]{{\color{blue} #1}}

\renewcommand{\paragraph}[1]{\vspace{-0.5ex}\textbf{#1}}

\newcommand{\titl}{Constant-Time Predictive Distributions for Gaussian Processes}
\newcommand{\titlshort}{Constant-Time Predictive Distributions for Gaussian Processes}
\newcommand{\authorinfo}{
  \icmlsetsymbol{equal}{*}
  \begin{icmlauthorlist}
    \icmlauthor{Geoff Pleiss}{cornell}
    \icmlauthor{Jacob R. Gardner}{cornell}
    \icmlauthor{Kilian Q. Weinberger}{cornell}
    \icmlauthor{Andrew Gordon Wilson}{cornell}
  \end{icmlauthorlist}

  \icmlaffiliation{cornell}{Cornell University}
  \icmlcorrespondingauthor{Geoff Pleiss}{geoff@cs.cornell.edu}
  \icmlcorrespondingauthor{Jacob R. Gardner}{jrg365@cornell.edu}
  \icmlcorrespondingauthor{Andrew Gordon Wilson}{andrew@cornell.edu}
  \icmlkeywords{Gaussian Processes}
}

\icmltitlerunning{\titlshort}
\begin{document}
  \twocolumn[
    \icmltitle{\titl}
    \authorinfo
    \vskip 0.3in
  ]
  \printAffiliationsAndNotice{}

  \begin{abstract}
	   One of the most compelling features of Gaussian process (GP) regression is its ability to provide well-calibrated posterior distributions.
	   Recent advances in inducing point methods have sped up GP marginal likelihood and posterior mean computations,
     leaving posterior covariance estimation and sampling as the remaining computational bottlenecks.
     In this paper we address these shortcomings by using the Lanczos algorithm to rapidly approximate the predictive covariance matrix.
     Our approach, which we refer to as \methodname{} (LanczOs Variance Estimates), substantially improves time and space complexity.
     In our experiments, LOVE computes covariances up to $2,\!000$ times faster and draws samples $18,\!000$ times faster than existing methods, all \emph{without} sacrificing accuracy.
  \end{abstract}

  %!TEX root=../main.tex
\section{Introduction}
Gaussian processes (GPs) are fully probabilistic models which can naturally estimate predictive uncertainty through posterior variances.
These uncertainties play a pivotal role in many application domains.
For example, uncertainty information is crucial when incorrect predictions could have catastrophic consequences, such as in medicine \cite{schulam2017if} or large-scale robotics \cite{deisenroth2015gaussian};
Bayesian optimization approaches typically incorporate model uncertainty when choosing actions \cite{snoek2012practical,deisenroth2011pilco,wang2017max};
and reliable uncertainty estimates are arguably useful for establishing trust in predictive models,
especially when predictions would be otherwise difficult to interpret
\cite{doshi2017roadmap,zhou2017effects}.

\emph{Although predictive uncertainties are a primary advantage of GP models, they have recently become their computational bottleneck.}
Historically, the use of GPs has been limited to problems with small datasets, since learning and inference computations na\"ively scale cubically with the number of data points ($n$).
Recent advances in \emph{inducing point methods} have managed to scale up GP training and computing predictive means to larger datasets \cite{snelson2006sparse,quinonero2005unifying,titsias2009variational}.
\emph{Kernel Interpolation for Scalable Structured GPs} (KISS-GP) is one such method that scales to millions of data points \cite{wilson2015kernel,wilson2015thoughts}.
For a test point $\x^*$, KISS-GP expresses the predictive mean as ${\blue\ba^\top} \bw(\x^*) $, where $\blue \ba$ is a pre-computed vector dependent only on training data, and $\bw(\x^*)$ is a sparse interpolation vector.
This formulation affords the ability to compute predictive means in \emph{constant time}, independent of $n$.

\begin{figure}[t!]
  \centering
  \includegraphics[width=0.8\columnwidth]{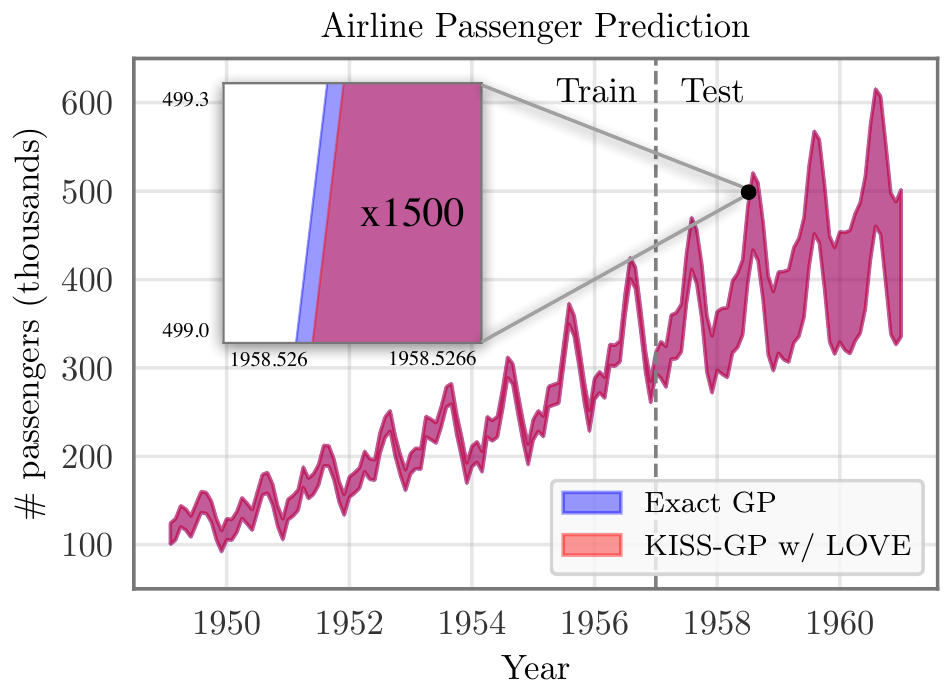}
  \caption{
    Comparison of predictive variances on airline passenger extrapolation.
    The variances predicted with \methodname{} are accurate within $10^{-4}$, yet can be computed orders of magnitude faster.
    \label{fig:airline_results}
  }
\end{figure}

However, these computational savings do not extend naturally to predictive uncertainties.
With KISS-GP, computing the predictive covariance between two points requires $\bigo{n + m \log m}$ computations, where $m$ is the number of inducing points (see \autoref{tab:running_times}).
While this asymptotic complexity is lower than standard GP inference and alternative scalable approaches, it becomes prohibitive when $n$ is large, or when making many repeated computations.
Additionally, drawing samples from the predictive distributions -- a necessary operation in many applications -- is similarly expensive.
Existing fast approximations for these operations \cite{papandreou2011efficient,wilson2015thoughts,wang2017max} typically incur a significant amount of error.
Matching the reduced complexity of predictive mean inference without sacrificing accuracy has remained an open problem.

\begin{table*}[t!]
  \caption{
    Asymptotic complexities of predictive (co)variances ($n$ training points, $m$ inducing points, $k$ Lanczos/CG iterations)
    and sampling from the predictive distribution ($s$ samples, $t$ test points).
    \label{tab:running_times}
  }
  \vspace{0.5ex}
  \centering
  \resizebox{\textwidth}{!}{%
    \begin{tabular}{ |c||c|c||c||c| }
      \hline
      \multirow{2}*{\bf Method} & \multicolumn{2}{c||}{\bf Pre-computation}  & {\bf Computing variances} & {\bf Drawing $s$ samples} \\
                                & \multicolumn{1}{c}{(time)} & (storage) & (time) & (time) \\
      \hhline{|=#=|=|=#=|}
      Standard GP
      & $\bigo{n^3}$
      & $\bigo{n^2}$
      & $\bigo{n^2}$
      & $\bigo{t n^2 + t^2 (n + s) + t^3}$
      \\
      SGPR
      & $\bigo{nm^2}$
      & $\bigo{m^2}$
      & $\bigo{m^2}$
      & $\bigo{t m^2 + t^2 (m + s) + t^3}$
      \\
      KISS-GP
      & --
      & --
      & $\bigo{k (n \! + \! m \log m)}$
      & $\bigo{k t (n \! + \! m \log m) \! + \! t^2 (m + s)  \! + \! t^3}$
      \\ \hline
      {\color{\ourmethodcolor} KISS-GP (w/ \methodname{})}
      & {\color{\ourmethodcolor}$\bigo{k (n \! + \! m \log m)}$}
      & {\color{\ourmethodcolor}$\bigo{km}$}
      & {\color{\ourmethodcolor}$\bigo{k}$}
      & {\color{\ourmethodcolor} $\bigo{k s (t + m)}$} \\
      \hline
    \end{tabular}
  }
  \vspace{-2ex}
\end{table*}

In this paper, we provide a solution based on the tridiagonalization algorithm of \citet{lanczos1950iteration}.
Our method takes inspiration from KISS-GP's mean computations: we express the predictive covariance between $\x^*_i$ and $\x^{*}_j$ as
$\bw(\x^*_i)^\top \blue{C} \: \bw(\x^{*}_j)$,
where $\blue C$ is an $m \times m$ matrix dependent only on training data.
However, we take advantage of the fact that $\blue C$ affords fast matrix-vector multiplications (MVMs) and avoid explicitly computing the matrix.
Using the Lanczos algorithm, we can efficiently decompose $\blue C$ as two rank-$k$ matrices $\blue C\!\approx\! \blue{R^T R^\prime}$ in \emph{nearly linear time}.
After this one-time upfront computation, and due to the special structure of $\blue {R,R^\prime}$, all variances can be computed in \emph{constant time} -- $\bigo{k}$ -- per (co)variance entry.
We extend this method to sample from the predictive distribution at $t$ points in $\bigo{t + m}$ time -- independent of training dataset size.

We refer to this method as LanczOs Variance Estimates, or \methodname{} for short.\footnote{
  \methodname{} is implemented in the GPyTorch library.
  Examples are available at \url{http://bit.ly/gpytorch-examples}.
}
\methodname{} has the lowest asymptotic complexity for computing predictive (co)variances and drawing samples with GPs.
We empirically validate \methodname{} on seven datasets and find that it consistently provides substantial speedups over existing methods \emph{without sacrificing accuracy}.
Variances and samples are accurate to within four decimals, and can be computed \emph{up to 18,000 times faster.}

  %!TEX root=../main.tex
\section{Background}
A Gaussian process is a prior over \emph{functions}, $p(f(\mathbf{x}))$, specified by a \emph{prior mean function} $\mu(\x)$ and \emph{prior covariance function} $k(\x, \x')$. Given a dataset of observations $\mathcal{D} = (X, \y) = \{(\x_i, y_i)\}_{i=1}^{n}$ and a Gaussian noise model, the posterior $p(f(\x) \posterior)$ is again a Gaussian process with mean $\mu_{f \mid \dset}(\x^*)$ and covariance $k_{f \mid \dset}(\x^*,\x^{*\prime})$:
\begin{align}
  \mu_{f \mid \dset}(\x^*) &= \mu(\x^*) + \bk_{\X \x^*}^\top  \widehat K_{\X\!\X}^{-1} (\y - \mu(\X)) \label{eq:pred_mean}, \\
  k_{f\mid\dset}(\x^*, \x^{*\prime}) &= k_{\x^{*} \x^{*\prime}} - \bk_{\X \x^{*}}^\top \widehat K_{\X\!\X}^{-1} \bk_{\X \x^{*\prime}},
    \label{eq:pred_covar}
\end{align}
where $K_{\A\B}$ denotes the kernel matrix between $\A$ and $\B$, $\widehat{K}_{\X\X} = K_{\X\X} + \sigma^2 I$ (for observed noise $\sigma$) and $\y = [y(\x_1),\dots,y(\x_n)]^{\top}$.
Given a set of $t$ test points $\X^{*}$, the equations above give rise to a $t$ dimensional multivariate Gaussian joint distribution
$p([f(\x^{*}_{1}),...,f(\x^{*}_{t}))]\posterior)$ over the function values of the $t$ test points. This last property allows for sampling functions from a posterior Gaussian process by sampling from this joint predictive distribution. For a full overview, see \cite{rasmussen2006gaussian}.

\subsection{Inference with matrix-vector multiplies}
Computing predictive means and variances with \eqref{eq:pred_mean} and \eqref{eq:pred_covar} requires computing solves with the kernel matrix $\widehat K_{X \! X}$ (e.g. $\widehat K_{X \! X}^{-1} \mathbf y$).
These solves are often computed using the Cholesky decomposition of $\widehat K_{\X \! \X} = L L^\top$, which requires $\bigo{n^3}$ time to compute.
{Linear conjugate gradients} (CG) provides an alternative approach, computing solves through matrix-vector multiplies (MVMs).
CG exploits the fact that the solution $A^{-1}\bb$ is the unique minimizer of the quadratic function $f(\x) =\frac{1}{2}\x^{\top}A\x - \x^{\top} \bb$ for positive definite matrices \cite{golub2012matrix}.
This function is minimized with a simple three-term recurrence, where each iteration involves a single MVM with the matrix $A$.

After $n$ iterations CG is guaranteed to converge to the exact solution $A^{-1} \bb$, although in practice numerical convergence may require substantially fewer than $n$ iterations.
Extremely accurate solutions typically require only $k \ll n$ iterations (depending on the conditioning of $A$) and $k\leq 100$ suffices in most cases~\cite{golub2012matrix}.
For the kernel matrix $\widehat K_{X \! X}$, the standard running time of $k$ CG iterations is $\bigo{k n^2}$ (the time for $k$ MVMs).
This runtime, which is already faster than the Cholesky decomposition, can be greatly improved if the kernel matrix $K_{\X\!\X}$ affords fast MVMs.
Fast MVMs are possible if the data are structured \cite{cunningham2008fast,saatcci2012scalable}, or by using a structured inducing point method \cite{wilson2015kernel}.

\subsection{The Lanczos algorithm}
\label{subsec:lanczos}
The Lanczos algorithm factorizes a symmetric matrix $A \in \reals^{n \times n}$ as $QTQ^\top$, where $T \! \in \! \reals^{n \times n}$ is symmetric tridiagonal and $Q \! \in \! \reals^{n \times n}$ is orthonormal. For a full discussion of the Lanczos algorithm see \citet{golub2012matrix}.
Briefly, the Lanczos algorithm uses a probe vector $\bb$ and computes an orthogonal basis of the Krylov subspace $\mathcal{K} (A, \bb)$:
\[
  \mathcal{K}(A,\bb) = \text{span}\left\{ \bb, A \bb, A^2 \bb, \ldots, A^{n-1} \bb \right\}.
\]
Applying Gram-Schmidt orthogonalization to these vectors produces the columns of $Q$, $\left[ {\bb}/{\Vert \bb \Vert}, \bq_2, \bq_3, \ldots, \bq_n \right]$ (here $\Vert \bb \Vert$ is the Euclidean norm of $\bb$).
The orthogonalization coefficients are collected into $T$.
Because $A$ is symmetric, each vector needs only be orthogonalized against the two preceding vectors, which results in the tridiagonal structure of $T$ \cite{golub2012matrix}.
The orthogonalized vectors and coefficients are computed in an iterative manner.
$k$ iterations produces the first $k$ orthogonal vectors of $Q_k = \left[ \bq_1, \ldots, \bq_k \right] \! \in \! \reals^{n \times k}$)
and their corresponding coefficients $T_k \! \in \! \reals^{k \times k}$.
Similarly to CG, these $k$ iterations require only $\bigo{k}$ matrix vector multiplies with the original matrix $A$, which again is ideal for matrices that afford fast MVMs.

The Lanczos algorithm can be used in the context of GPs for computing log determinants \cite{dong2017scalable},
and can be used to speed up inference when there is product structure \cite{gardner2018product}.
Another application of the Lanczos algorithm is performing matrix solves \cite{lanczos1950iteration,parlett1980new,saad1987lanczos}.
Given a symmetric matrix $A$ and a single vector $\bb$, the matrix solve $A^{-1} \bb$ is computed by starting the Lanczos algorithm of $A$ with probe vector $\bb$.
After $k$ iterations, the solution $A^{-1} \bb$ can be approximated using the computed Lanczos factors $Q_{k}$ and $T_{k}$ as
\begin{equation}
  A^{-1} \bb \approx \Vert \bb \Vert Q_k T_k^{-1} \be_1,
  \label{eq:lanczos_single_solve}
\end{equation}
where $\be_1$ is the unit vector $[1, 0, 0, \ldots, 0]$.
%In fact, the linear CG algorithm can be derived from \eqref{eq:lanczos_single_solve} when $A$ is positive definite \cite{golub2012matrix}.
These solves tend to be very accurate after $k \ll n$ iterations, since the eigenvalues of the $T$ matrix converge rapidly to the largest and smallest eigenvalues of $A$ \cite{demmel1997applied}.
The exact rate of convergence depends on the conditioning of $A$ \cite{golub2012matrix}, although in practice we find that $k\leq100$ produces extremely accurate solves for most matrices (see \autoref{sec:results}).
In practice, CG tends to be preferred for matrix solves since Lanczos solves require storing the $Q_k \! \in \! \reals^{n \times k}$ matrix.
However, one advantage of Lanczos is that the $Q_k$ and $T_k$ matrices can be used to jump-start subsequent solves $A^{-1} \bb'$.
\citet{parlett1980new}, \citet{saad1987lanczos}, \citet{schneider2001krylov}, and \citet{nickisch2009bayesian} argue that solves can be approximated as
\begin{equation}
  A^{-1} \bb' \approx Q_k T_k^{-1} Q_k^\top \bb',
  \label{eq:lanczos_solve}
\end{equation}
where $Q_k$ and $T_k$ come from a previous solve $A^{-1} \bb$.

\subsection{Kernel Interpolation for Scalable Structured GPs}
Structured kernel interpolation (SKI) \cite{wilson2015kernel} is an inducing point method explicitly designed for the MVM-based inference described above.
Given a set of $m$ inducing points, $U = [\bu_1, \ldots, \bu_m]$, SKI assumes that a data point $\x$ is well-approximated as a \emph{local interpolation} of $U$.
Using cubic interpolation \cite{keys1981cubic}, $\x$ is expressed in terms of its 4 closest inducing points, and the interpolation weights are captured in a sparse vector $\bw_\x$.
The $\bw_\x$ vectors are used to approximate the kernel matrix $K_{\X\X} \approx \tilde K_{\X\X}$:
\begin{equation}
  \tilde K_{\X \X} = \W_{\X}^\top \K_{UU} \W_{\X}.
  \label{eq:ski}
\end{equation}
Here, $W_{\X} = [\bw_{\x_1}, \ldots, \bw_{\x_n}]$ contains the interpolation vectors for all $\x_i$, and $K_{UU}$ is the covariance matrix between inducing points.
MVMs with $\tilde K_{XX}$ (i.e. $W_X^\top K_{UU} W_X \bv$) require at most $\bigo{n + m^2}$ time due to the $\bigo{n}$ sparsity of $W_X$.
\citet{wilson2015kernel} reduce this runtime even further with \emph{Kernel Interpolation for Scalable Structured GPs} (KISS-GP),
%an instantiation of their SKI approach
in which all inducing points $U$ lie on a regularly spaced grid.
This gives $K_{UU}$ Toeplitz structure (or Kronecker and Toeplitz structure),
resulting in the ability to perform MVMs in at most $\bigo{n + m \log m}$ time.

\paragraph{Computing predictive means.}
One advantage of KISS-GP's fast MVMs is the ability to perform constant time predictive mean calculations \cite{wilson2015thoughts}.
Substituting the KISS-GP approximate kernel into \eqref{eq:pred_mean} and assuming a prior mean of 0 for notational brevity, the predictive mean is given by
\begin{equation}
  \mu_{f \mid \dset}(\x^{*}) = \bw_{\x^{*}}^\top \blue{\underbrace{\K_{U\!U}\W_{\X}(\W_{\X}^{\top}\K_{U\!U}\W_{\X} \! + \! \sigma^{2} \! I)^{\!-1}\y}_{\ba}}.
  \label{eq:pred_mean_ski}
\end{equation}
Because $\bw_{\x^{*}}$ is the only term in \eqref{eq:pred_mean_ski} that depends on $\x^{*}$, the remainder of the equation (denoted as $\blue\ba$) can be pre-computed: $\mu_{f \mid \dset}(\x^{*}) = \bw_{\x^{*}}^\top \blue \ba$.
(Throughout this paper {\color{blue} blue} highlights computations that don't depend on test data.)
This pre-computation takes $\bigo{n + m\log m}$ time using CG.
After computing $\blue \ba$, the multiplication $\bw_{\x^{*}}^\top \blue \ba$ requires $\bigo{1}$ time, as $\bw_{\x^{*}}$ has only four nonzero elements.

  %!TEX root=../main.tex
\section{Lanczos Variance Estimates (\href{https://www.youtube.com/watch?v=t5ze_e4R9QY}{\methodname{}})}
\label{sec:method}

In this section we introduce \methodname{}, an approach to efficiently approximate the predictive covariance
\begin{align*}
  k_{f\mid\dset}(\x^*_i, \x^*_j) = k_{\x^{*}_i \x^{*}_j} - \bk_{\X \! \x^{*}_i}^\top (K_{\X\!\X} + \sigma^2 \eye)^{-1} \bk_{\X \! \x^{*}_j},
\end{align*}
where $\X^{*} = [ \x^{*}_1, \ldots \x^{*}_t ]$ denotes a set of $t$ test points, and $\X = [\x_1, \ldots \x_n]$ a set of $n$ training points.
Solving $(K_{\X \X} + \sigma^2 \eye)^{-1}$ na\"ively takes $\bigo{n^3}$ time, but is typically performed as a one-type pre-computation during training since $K_{\X \X}$ does not depend on the test data.
At test time, all covariance calculations will then cost $\bigo{n^2}$.
In what follows, we build a significantly faster method for (co-)variance computations, mirroring this \emph{pre-compute phase} and \emph{test phase} structure.

\subsection{A first $\bigo{m^2}$ approach with KISS-GP}
It is possible to obtain some computational savings when using inducing point methods.
For example, we can replace $\bk_{X \! \x^*_i}$ and $K_{X \! X}$ with their corresponding KISS-GP approximations,
$\tilde{\bk}_{X \! \x^*_i}$ and $\tilde{K}_{X \! X}$, which we restate here:
\begin{align*}
  \tilde{\bk}_{\X \x^*_i} = W^\top_{\X} \K_{U \! U} \bw_{\x^*_i},
  \: \: \: & \: \: \:
  \tilde{K}_{\X \X} = W^\top_{\X} \K_{U \! U} W_{\X}.
\end{align*}
$W_\X$ is the sparse interpolation matrix for training points $X$.
$\bw_{\x^{*}_i}$ and $\bw_{\x^{*}_j}$ are the sparse interpolations for $\x^{*}_i$ and $\x^{*}_j$ respectively.
Substituting these into \eqref{eq:pred_covar} results in the following approximation to the predictive covariance:
\begin{align}
  k_{\!f\mid\dset}(\x^*_i, \x^*_j) &\approx k_{\x^{*}_i \! \x^{*}_j} - \tilde{\bk}_{\X \! \x^{*}_i}^\top (\tilde{K}_{\X\!\X} + \sigma^2 \eye)^{-1} \tilde{\bk}_{\X \! \x^{*}_j}.
    \label{eq:pred_covar_approx}
\end{align}
By fully expanding the second term in \eqref{eq:pred_covar_approx}, we obtain
\begin{equation}
  \bw_{\x^*_i}^{\top} {\color{blue}\underbrace{{\K_{U \! U} W_{\X} (\tilde{K}_{\X\!\X} + \sigma^2 \eye)^{-1} W^\top_{\X} \K_{U \! U}}}_{\blue C}} \bw_{\x^*_j}
  \label{eq:pred_covar_ski_cache}
\end{equation}

\paragraph{Precomputation phase.}
$\blue C$, the braced portion of \eqref{eq:pred_covar_ski_cache}, does not depend on the test points $\x^{*}_{i}$, $\x^{*}_{j}$ and therefore can be pre-computed during training.
The covariance becomes:
\begin{align}
  k_{f\mid\dset}(\x^*_i, \x^*_j) &\approx k_{\x^{*}_i \x^{*}_j} - \bw_{\x^{*}_i}^\top \: \blue C \: \bw_{\x^{*}_j}
    \label{eq:pred_covar_ski_naive}
\end{align}
In \eqref{eq:pred_covar_ski_cache}, we see that primary cost of computing $\blue C$ is the $m$ solves with $\tilde{K}_{\X\!\X} + \sigma^2 \eye$: one for each column vector in $W^{\top}_{\X} \K_{U \! U}$.
Since $\tilde K_{\X\!\X}$ is a KISS-GP approximation, each solve is $\bigo{n + m\log m}$ time with CG \cite{wilson2015kernel}.
The total time for $m$ solves is $\bigo{mn + m^2 \log m}$.
%Computing the right hand sides takes $\bigo{nm}$ time total because $W^{\top}_{\X}$ is an $n \times m$ matrix with four non-zero elements per row.

\paragraph{Test phase.}
%In the \emph{test phase} we can compute individual predictive variances in $\bigo{m}$ time.
As $\bw^{*}_{i}$ contains only four nonzero elements,
the inner product of $\bw^{*}_{i}$ with a vector takes $\bigo{1}$ time,
and the multiplication $\bw^{*\top}_{i}\blue C$ requires $\bigo{m}$ time during testing.

\paragraph{Limitations.}
Although this technique offers computational savings over non-inducing point methods, the quadratic dependence on $m$ in the pre-computation phase is a computational bottleneck.
In contrast, all other operations with KISS-GP require at most linear storage and near-linear time.
Indeed, one of the hallmarks of KISS-GP is the ability to use a very large number of inducing points $m = \Theta(n)$ so that kernel computations are nearly exact.
Therefore, in practice a quadratic dependence on $m$ is infeasible and so no terms are pre-computed.\footnote{
  \citet{wilson2015thoughts} (Section 5.1.2) describe an alternative procedure that approximates $\blue C$ as a diagonal matrix
  for fast variances, but typically incurs much greater (e.g., more than 20\%) error, which is dominated by the number of terms in
  a stochastic expansion, compared to the number of inducing points.}
Variances are instead computed using \eqref{eq:pred_covar_approx}, computing each term in the equation from scratch.
Using CG, this has a cost of $\bigo{k n + k m \log m}$ for \emph{each} (co-)variance computation, where $k$ is the number of CG iterations.
This dependence on $n$ and $m$ may be cumbersome when performing many variance computations.

\subsection{Fast predictive (co-)variances with \methodname{}}
We propose to overcome these limitations through an altered pre-computation step.
In particular, we approximate $\blue C$ in \eqref{eq:pred_covar_ski_naive} as a low rank matrix.
Letting $\blue R$ and $\blue{R'}$ be $k \times m$ matrices such that $\blue R^\top \blue{R'} \approx \blue C$, we rewrite \eqref{eq:pred_covar_ski_naive} as:
%
% \begin{align}
%   k_{f\mid\dset}(\x^*_i, \x^*_j) &\approx k_{\x^{*}_i \x^{*}_j} - \bw_{\x^{*^\top}_i} \blue R^{\top}\blue{R'} \: \bw_{\x^{*}_j}.\nonumber \\
%                                  &\approx k_{\x^{*}_i \x^{*}_j} - (\blue R \bw_{\x^{*}_i})^\top (\blue{R'} \: \bw_{\x^{*}_j}).
%     \label{eq:pred_covar_ski_fast}
% \end{align}
\begin{align}
  k_{f\mid\dset}(\x^*_i, \x^*_j) &\approx k_{\x^{*}_i \x^{*}_j} - (\blue R\bw^{*}_i )^{\top}\blue{R'} \: \bw^{*}_j.
    \label{eq:pred_covar_ski_fast}
\end{align}
Variance computations with \eqref{eq:pred_covar_ski_fast} take $\bigo{k}$ time
due to the sparsity of $\bw_{\x^{*}_i}$ and $\bw_{\x^{*}_j}$.
Recalling the Lanczos approximation
$
    (\tilde{K}_{\X\X} + \sigma^2 \eye)^{-1}\bb \approx Q_{k}T^{-1}_{k}Q_{k}^{\top}\bb
$
from \autoref{subsec:lanczos}, we can efficiently derive forms for $\blue R$ and $\blue{R'}$:
\begin{align*}
  \blue C &= \K_{U \! U} W_{\X} \underbrace{(\tilde{K}_{\X\!\X} + \sigma^2 \eye)^{-1}}_{\text{Apply Lanczos}} W^\top_{\X} \K_{U \! U} \\
         &\approx K_{UU} W_X (Q_k T_k^{-1} Q_k^\top) W_X^\top K_{UU} \\
         &= \underbrace{K_{UU} W_X Q_k}_{\blue R^\top} \underbrace{T_k^{-1} Q_k^\top W_X^\top K_{UU}}_{\blue{R'}}
\end{align*}
To compute $\blue R$ and $\blue{R'}$, we perform $k$ iterations of Lanczos to achieve $(\tilde{K}_{\X\X} + \sigma^2\eye)  \approx Q_{k}T_{k}Q_{k}^{\top}$ using the average column vector $\mathbf{b}=\frac{1}{m}W_X^\top K_{UU}\mathbf{1}$ as a probe vector.
This partial Lanczos decomposition requires $k$ MVMs with $(\tilde K_{XX} + \sigma^2 \eye)$ for a total of $\bigo{k n + k m \log m}$ time (because of the KISS-GP approximation).
$\blue R$ and $\blue{R'}$ are $m \times k$ matrices, and thus require $\bigo{mk}$ storage.

\paragraph{To compute $\blue R$,} we first multiply $W_{\X}Q_{k}$, which takes $\bigo{kn}$ time due sparsity of $W_\X$.
The result is a $m \times k$ matrix.
Since $\K_{UU}$ has Toeplitz structure, the multiplication $\K_{U\!U} (W_{\X}Q_{k})$ takes $\bigo{km \log m}$ time \cite{saatcci2012scalable}.
Therefore, computing $\blue R$ takes $\bigo{kn + km\log m}$ total time.

\paragraph{To compute $\blue {R'}$,} note that $\blue{R'} = T^{-1}_{k}\blue R^{\top}$.
Since $\tilde K_{X\!X}$ is positive definite, $T$ is as well (by properties of the Lanczos algorithm).
We thus compute $T^{-1}_k \blue R^\top$ using the Cholesky decomposition of $T$.
Computing/using this decomposition takes $\bigo{km}$ time since $T$ is tridiagonal \cite{loan1999introduction}.

In total, the entire pre-computation phase takes only $\bigo{kn + km \log m}$ time.
This is the same amount of time of a single marginal likelihood computation.
We perform the pre-computation as part of the training procedure since none of the terms depend on test data.
Therefore, during test time all predictive variances can be computed in $\bigo{k}$ time using \eqref{eq:pred_covar_ski_fast}.
As stated in \autoref{subsec:lanczos}, $k$ depends on the conditioning of the matrix and not its size \cite{golub2012matrix}.
$k\leq100$ is sufficient for most matrices in practice \cite{golub2012matrix}, and therefore $k$ can be considered to be constant.
Running times are summarized in \autoref{tab:running_times}.

\subsection{Predictive distribution sampling with \methodname{}}
\label{sec:sampling_method}

\methodname{} can also be used to compute predictive \emph{covariances} and operations involving the predictive covariance matrix.
Let $X^* = [\x^*_1, \ldots, \x^*_t]$ be a test set of $t$ points.
%In many applications \eqref{eq:pred_covar} will be used only to compute the predictive variance terms for each $\x^*_i$,  i.e. $k_{f \mid \dset} (\x^*_i, \x^*_i)$.
To draw samples from $\fn^* \! \mid \! \dset$ --- the predictive function evaluated on $\x^*_1, \ldots, \x^*_t$, the cross-covariance terms (i.e. $k_{f \mid \dset} (\x^*_i, \x^*_j)$) are necessary in addition to the variance terms ($k_{f \mid \dset} (\x^*_i, \x^*_i)$).
Sampling GP posterior functions is a common operation.
In Bayesian optimization for example, several popular acquisition functions -- such as predictive entropy search \cite{hernandez2014predictive}, max-value entropy search \cite{wang2017max}, and knowledge gradient \cite{frazier2009knowledge} -- require posterior sampling.

However, posterior sampling is an expensive operation when querying at many test points.
The predictive distribution $\fn^* \! \mid \! \dset$ is multivariate normal with mean $\mu_{f \mid \dset} (\X^*) \in \reals^t$ and covariance $k_{f \mid \dset} (\X^*, \X^*) \in \reals^{t \times t}$.
We sample $\fn^* \! \mid \! \dset$ by reparametrizing Gaussian noise samples $\mathbf{v} \sim \normaldist{0}{\eye{}^{t\!\times \!t}}$:
\begin{equation}
  \mu_{f \mid \dset} (\X^*) + S \mathbf{v},
  \label{eq:sample}
\end{equation}
where $S$ is a matrix such that $S S^\top = k_{f \mid \dset} (\X^*, \X^*)$.
Typically $S S^\top$ is taken to be the Cholesky decomposition of $k_{f \mid \dset} (\X^*, \X^*)$.
Computing this decomposition incurs a $\bigo{t^3}$ cost on top of the $\bigo{t^2}$ covariance evaluations.
This may be costly, even with constant-time covariance computations.
Parametric approximations are often used instead of exact sampling \cite{deisenroth2011pilco}.

\paragraph{A Fast Low-Rank Sampling Matrix.} We use \methodname{} and KISS-GP to rewrite \eqref{eq:pred_covar_ski_fast} as
\begin{align}
  k_{f \mid \dset} (\X^*, \X^*)
  &\approx W^\top_{X^*} \blue{K_{UU}} W_{X^*} - (\blue R W_{X^*})^\top (\blue{R'} W_{X^*})
    \notag \\
    &= W^\top_{X^*} \blue{\left( K_{UU} - R^\top R' \right)} W_{X^*}.
    \label{eq:pred_covar_ski_interp_form12}
\end{align}
where $W_{X^*} = [\bw_{x^*_1}, \ldots, \bw_{x^*_n}]$ is the interpolation matrix for test points.
We have reduced the full covariance matrix to a test-independent term ($\blue{ K_{UU} - R^\top R' }$) that can be pre-computed.
We apply the Lanczos algorithm on this term during pre-computation to obtain a rank-$k$ approximation:
\begin{align}
  \blue{K_{UU} - R^\top R' \approx Q'_k T'_k Q_k^{\prime\top} }.
  \label{eqn: lancapprox}
\end{align}
This Lanczos decomposition requires $k$ matrix vector multiplies with $\blue{ K_{UU} - R^{\top}R' }$, each of which requires $\bigo{m \log m}$ time.
Substituting \eqref{eqn: lancapprox} into \eqref{eq:pred_covar_ski_interp_form12}, we get:
\begin{align}
  k_{f \mid \dset} (\X^*, \X^*) = W^\top_{X^*} \blue{Q'_{k}T'_{k}Q_{k}^{\prime\top}} W_{X^*}.
  \label{eq:sampling_pre_cholesky}
\end{align}
If we take the Cholesky decomposition of $T'_k = \!L L^\top$ (a $\bigo{k}$ operation since $T'_{k}$ is tridiagonal), we rewrite \eqref{eq:sampling_pre_cholesky} as
\begin{align}
  k_{f \mid \dset} (\X^*, \X^*)
  &\approx  W^\top_{X^*} \blue{\underbrace{Q'_k L}_{S}} \blue{\underbrace{L^\top Q_k^{\prime\top}}_{S^\top}} W_{X^*}.
    \label{eq:pred_covar_ski_interp_form}
\end{align}
Setting $\blue{ S= Q'_k L_{T_k} }$, we see that $k_{f \mid \dset} (\X^*, \X^*) = (W_X^\top \blue S)(W_X^\top \blue S)^{\top}$.
$\blue S \! \in \! \reals^{m \times k}$ can be precomputed and cached since it does not depend on test data.
In total, this pre-computation takes $\bigo{k m \log m + m k^2}$ time in addition to what is required for fast variances.
To sample from the predictive distribution, we need to evaluate \eqref{eq:sample}, which involves multiplying $W^\top_{X^*} S \mathbf{v}$.
Multiplying $\mathbf{v}$ by $S$ requires $\bigo{mk}$ time, and finally multiplying by $W^{\top}_{X^{*}}$ takes $\bigo{tk}$ time.
Therefore, drawing $s$ samples (corresponding to $s$ different values of $\mathbf{v}$) takes $\bigo{sk(t + m)}$ time total during the testing phase (see \autoref{tab:running_times}) -- a \emph{linear} dependence on $t$.

\begin{algorithm2e}[t]
  \SetAlgoLined
  \SetKwInOut{Input}{Input}
  \SetKwInOut{Output}{Output}
  \newlength\inputlen
  \newcommand\NextInput[1]{%
    \settowidth\inputlen{\Input{}}%
    \setlength\hangindent{1.5\inputlen}%
    \hspace*{\inputlen}#1\\
  }
  \newcommand\graycomment[1]{\footnotesize\ttfamily\textcolor{gray}{#1}}
  \SetCommentSty{graycomment}
  \SetKwFunction{mvmkxx}{mvm\_K\textsubscript{XX}}
  \SetKwFunction{mvmkux}{mvm\_K\textsubscript{UX}}
  \SetKwFunction{lanczos}{lanczos\textsubscript{k}}
  \SetKwFunction{sparsemm}{sparse\_mm}
  \SetKwFunction{choleskyfactor}{cholesky\_factor}
  \SetKwFunction{choleskysolve}{cholesky\_solve}
  \caption{\methodname{} for fast predictive variances.}
  \label{alg:fast_pred_var}
    \Input{$\bw_{x^*_i}$, $\bw_{x^*_j}$ -- interpolation vectors for $\x^*_i$, $\x^*_j$}
    \NextInput{$k_{\x^{*}_i, \x^{*}_j}$ -- prior covariance between $\x^*_i$, $\x^*_j$}
    \NextInput{$\bb=\frac{1}{m}W_X^\top K_{UU}\mathbf{1}$ -- average col. of $W_X^\top K_{UU}$}
    \NextInput{\mvmkxx{}: func. that performs MVMs with $(W^\top_{X} K_{UU} W_X + \sigma^2 \eye) \approx \widehat K_{X\!X}$}
    \NextInput{\mvmkux{}: func. that performs MVMs with $(K_{UU} W_{X}) \approx K_{U\!X}$}
    \Output{Approximate predictive variance $k_{f \mid \mathcal D} ( \x^{*}_i, \x^{*}_j )$.}
    \BlankLine
    \If{$\blue R, \blue{R'}$ have not previously been computed}{
      $Q_k, T_k$ $\leftarrow$ \lanczos{ \mvmkxx, $\bb$ } \\
      \qquad \tcp{k iter. of Lanczos w/}
      \qquad \tcp{matrix $\widehat K_{\X\X}$ and probe vec. $\!\!\!\bb$ }
      $L_{T_k}$ $\leftarrow$ \choleskyfactor{ $T_k$ } \\
      $\blue R$ $\leftarrow$ $($ \mvmkux{ $Q_k$ } $)^\top$
      \tcp*{$R = Q_k^\top W^\top_X K_{UU}$  }
      $\blue{R'}$ $\leftarrow$ \choleskysolve{$\blue R$, $L_{T_k}$} \\
      \qquad \tcp{$R' = T_k^{-1} Q_k^\top W^\top_X K_{UU}$  }
    }
    $\bu$ $\leftarrow$ \sparsemm{ $\blue R$, $\bw_{x^*_i}$ } \\
    $\bv$ $\leftarrow$ \sparsemm{ $\blue{R'}$, $\bw_{x^*_j}$ } \\
    \Return{ $k_{\x^{*}_i, \x^{*}_j} - \bu^T \bv$ } \\
\end{algorithm2e}

\subsection{Extension to additive kernel compositions}
\methodname{} is applicable even when the KISS-GP approximation is used with an additive composition of kernels,
\begin{equation}
  \tilde{k}(\x_i, \x_j) =
  \bw^{(1)\top}_{\x_i} K^{(1)}_{U \! U} \bw^{(1)}_{\x_j} + \ldots + \bw^{(d)\top}_{\x_i} K^{(d)}_{U \! U} \bw^{(d)}_{\x_j}.
  \notag
\end{equation}
Additive structure has recently been a focus in several Bayesian optimization settings, since the cumulative regret of additive models depends linearly on the number of dimensions
\cite{kandasamy2015high,wang2017batched,gardner2017discovering,wang2017max}.
Additionally, deep kernel learning GPs \citep{wilson2016stochastic,wilson2016deep} typically uses sums of one-dimensional kernel functions.
To apply \methodname{}, we note that additive composition can be re-written as
\begin{equation}
  \tilde{k}(\x_i, \x_j) =
  \begin{bmatrix}
    \bw^{(1)}_{\x_i} \\
    \vdots \\
    \bw^{(d)}_{\x_i}
  \end{bmatrix}^\top
  \!
  \begin{bmatrix}
    K^{(1)}_{U \! U} & \!\! \ldots \!\! & 0 \\
    \vdots & \!\! \ddots \!\! & \vdots \\
    0 & \!\! \ldots \!\! & K^{(d)}_{U \! U}
  \end{bmatrix}
  \!
  \begin{bmatrix}
    \bw^{(1)}_{\x_j} \\
    \vdots \\
    \bw^{(d)}_{\x_j}
  \end{bmatrix}.
  \label{eq:multi_dimensional_kernel_block}
\end{equation}
The block matrices in \eqref{eq:multi_dimensional_kernel_block} are analogs of their 1-dimensional counterparts in \eqref{eq:ski}.
Therefore, we can directly apply \autoref{alg:fast_pred_var}, replacing $W_X$, $\bw_{\x^*_i}$, $\bw_{\x^*_j}$, and $K_{UU}$ with their block forms.
The block $\bw$ vectors are $\bigo{d}$-sparse, and therefore interpolation takes $\bigo{d}$ time.
MVMs with the block $K_{UU}$ matrix take $\bigo{dm\log m}$ time by exploiting the block-diagonal structure. With $d$ additive components, predictive variance computations cost only a factor $\bigo{d}$ more than their 1-dimensional counterparts.

  %!TEX root=../main.tex
\section{Results}
\label{sec:results}
\begin{table*}[t!]
  \caption{
    Speedup and accuracy of KISS-GP/\methodname{} for predictive variances.
    KISS-GP and Exact GPs use deep kernel learning.
    Speed results are measured on GPUs.
    Accuracy is measured by Scaled Mean Average Error.
    ($n$ is the number of data, $d$ is the dimensionality.)
    \label{tab:large_dataset_results}
  }
  \vspace{0.5ex}
  \centering
  \resizebox{\textwidth}{!}{%
    %!TEX root=../main.tex
\begin{tabular}{ |ccc||c|c||c|c||c|c|c|c|| }
  \hline
  \multicolumn{3}{|c||}{\bf Dataset}
  & \multicolumn{2}{c||}{\thead{\bf Variance SMAE}}
  & \multicolumn{2}{c||}{\thead{\bf Speedup over KISS-GP (w/o \methodname{})}}
  & \multicolumn{4}{c||}{\thead{\bf Speedup over SGPR}}
  \\
  \cline{1-11}
  \multirow{2}{*}{Name}
  & \multirow{2}{*}{$n$}
  & \multirow{2}{*}{$d$}
  & \multirow{2}{*}{\thead{(vs KISS-GP w/o LOVE)}}
  & \multirow{2}{*}{\thead{(vs Exact GP)}}
  & \multirow{2}{*}{\thead{(from scratch)}}
  & \multirow{2}{*}{\thead{(after pre-comp.)}}
  & \thead{(from scratch)}
  & \thead{(after pre-comp.)}
  & \thead{(from scratch)}
  & \thead{(after pre-comp.)}
  \\
   &  &  &  &  &  &  & m=100 & m=100 & m=1000 & m=1000
  \\
  \hhline{|===#=|=#=|=#=|=|=|=#}
  \thead{\bf Airfoil}
  & $1,\!503$
  & $6$
  & $1.30 \times 10^{-5}$
  & $7.01 \times 10^{-5}$
  & $4 \times$
  & $84 \times$
  & $3 \times$
  & $49 \times$
  & $9 \times$
  & $183 \times$
  \\

  \thead{\bf Skillcraft}
  & $3,\!338$
  & $19$
  & $2.00 \times 10^{-7}$
  & $2.86 \times 10^{-4}$
  & $25 \times$
  & $167 \times$
  & $4 \times$
  & $70 \times$
  & $17 \times$
  & $110 \times$
  \\

  \thead{\bf Parkinsons}
  & $5,\!875$
  & $20$
  & $8.80 \times 10^{-5}$
  & $5.18 \times 10^{-3}$
  & $46 \times$
  & $443 \times$
  & $3 \times$
  & $33 \times$
  & $16 \times$
  & $152 \times$
  \\

  \thead{\bf PoleTele}
  & $15,\!000$
  & $26$
  & $2.90 \times 10^{-5}$
  & $1.08 \times 10^{-3}$
  & $78 \times$
  & $1178 \times$
  & $1.5 \times$
  & $40 \times$
  & $21 \times$
  & $343 \times$
  \\

  \thead{\bf Elevators}
  & $16,\!599$
  & $18$
  & $1.20 \times 10^{-6}$
  & --
  & $64 \times$
  & $1017 \times$
  & $2 \times$
  & $31 \times$
  & $20 \times$
  & $316 \times$
  \\

  \thead{\bf Kin40k}
  & $40,\!000$
  & $8$
  & $3.90 \times 10^{-7}$
  & --
  & $31 \times$
  & $2065 \times$
  & $8 \times$
  & $81 \times$
  & $12 \times$
  & $798 \times$
  \\

  \thead{\bf Protein}
  & $45,\!730$
  & $9$
  & $5.30 \times 10^{-5}$
  & --
  & $44 \times$
  & $1151 \times$
  & $10 \times$
  & $109 \times$
  & $20 \times$
  & $520 \times$
  \\
  \hline
\end{tabular}

  }
\end{table*}

In this section we demonstrate the effectiveness and speed of \methodname{}, both at computing predictive variances and also at posterior sampling.
%Our goal is to show that 1) \methodname{} produces uncertainties and samples that are indistinguishable from the state-of-the-art, and 2) that \methodname{} offers substantial speed improvements.
All \methodname{} variances are computed with $k=50$ Lanczos iterations,
and KISS-GP models use $m\!=\!10,\!000$ inducing points unless otherwise stated.
We perform experiments using the GPyTorch library.\footnote{
  \url{github.com/cornellius-gp/gpytorch}}
We optimize models with ADAM \cite{kingma2014adam} and a learning rate of $0.1$.
All timing experiments utilize GPU acceleration, performed on a NVIDIA GTX 1070.

\subsection{Predictive Variances}
\label{sec:results_variances}

\begin{table*}[t!]
  \caption{
    Accuracy and computation time of drawing samples from the predictive distribution.
    \label{tab:sampling_results}
  }
  \vspace{0.5ex}
  \centering
  \resizebox{\textwidth}{!}{%
    %!TEX root=../main.tex
\begin{tabular}{ |c||c|c|c|c|c||c|c|c|c|c| }
  \hline
  \multirow{4}{*}{\thead{\bf Dataset} }
  & \multicolumn{5}{c||}{\thead{\bf Sample Covariance Error} }
  & \multicolumn{5}{c|}{\thead{\bf Speedup over Exact GP w/ Cholesky} }

  \\
  \cline{2-11}

  & \thead{\bf Exact GP \\ \bf w/ Cholesky}
  & \thead{\bf Fourier \\ \bf Features}
  & \thead{\bf SGPR (m=100)}
  & \thead{\bf SGPR (m=1000)}
  & \thead{\bf KISS-GP \\ \bf w/ \methodname}
  & \thead{\bf Fourier \\ \bf Features}
  & \thead{\bf SGPR (m=100)}
  & \thead{\bf SGPR (m=1000)}
  & \thead{\bf KISS-GP \\ \bf w/ \methodname{} \\ (from scratch)}
  & \thead{\bf KISS-GP \\ \bf w/ \methodname{} \\ (after pre-comp.)}
  \\
  \hhline{|=#=|=|=|=|=#=|=|=|=|=|}

  \thead{\bf PolTele}
  & $\mathbf{8.8 \times 10^{-4}}$
  & $1.8 \times 10^{-3}$
  & $4.9 \times 10^{-3}$
  & $2.7 \times 10^{-3}$
  & $\mathbf{7.5 \times 10^{-4}}$
  & $22 \times$
  & $24 \times$
  & $3 \times$
  & $21 \times$
  & $\mathbf{881 \times}$
  \\

  \thead{\bf Elevators}
  & $\mathbf{2.6 \times 10^{-7}}$
  & $3.1 \times 10^{-4}$
  & $8.7 \times 10^{-6}$
  & $3.6 \times 10^{-6}$
  & $\mathbf{5.5 \times 10^{-7}}$
  & $31 \times$
  & $33 \times$
  & $4 \times$
  & $25 \times$
  & $\mathbf{1062 \times}$
  \\
  \hline

  \thead{\bf BayesOpt (Eggholder)}
  & $\mathbf{7.7 \times 10^{-4}}$
  & $1.5 \times 10^{-3}$
  & $\mathbf{8.1 \times 10^{-4}}$
  & --
  & $\mathbf{8.0 \times 10^{-5}}$
  & $16 \times$
  & $8 \times$
  & --
  & $19 \times$
  & $\mathbf{775 \times}$
  \\

  \thead{\bf BayesOpt (Styblinski-Tang)}
  & $\mathbf{5.4 \times 10^{-4}}$
  & $7.3 \times 10^{-3}$
  & $\mathbf{5.2 \times 10^{-4}}$
  & --
  & $\mathbf{5.2 \times 10^{-4}}$
  & $11 \times$
  & $8 \times$
  & --
  & $42 \times$
  & $\mathbf{18,\!100 \times}$
  \\
  \hline

\end{tabular}

  }
\end{table*}

We measure the accuracy and speed of KISS-GP/\methodname{} at computing predictive variances.
We compare variances computed with a KISS-GP/\methodname{} model against variances computed with an Exact GP ({\bf Exact}).
On datasets that are too large to run exact GP inference, we instead compare the KISS-GP/\methodname{} variances against KISS-GP variances computed in the standard way ({\bf KISS-GP w/o \methodname{}}).
\citet{wilson2015thoughts} show that KISS-GP variances recover the exact variance up to 4 decimal places.
Therefore, we will know that KISS-GP/\methodname{} produces accurate variances if it matches standard KISS-GP w/o \methodname{}.
We report the scaled mean absolute error (SMAE)\footnote{
  Mean absolute error divided by the variance of y.
} \cite{rasmussen2006gaussian} of \methodname{} variances compared against these baselines.
%(Note that the predictive posterior means are \emph{completely identical} with both methods, as \methodname{} only affects variance calculations.)
For each dataset, we optimize the hyperparameters of a KISS-GP model.
We then use the same hyperparameters for each baseline model when computing variances.

\paragraph{One-dimensional example.}
We first demonstrate \methodname{} on a complex one-dimensional regression task.
The airline passenger dataset ({\bf Airline}) measures the average monthly number of passengers from 1949 to 1961 \cite{hyndman2005time}.
We aim to extrapolate the numbers for the final 4 years (48 measurements) given data for the first 8 years (96 measurements).
%These data exhibit short-term periodic trends as well as long-term growth trends.  Consequentially,
Accurate extrapolation on this dataset requires a kernel function capable of expressing various patterns, such as the spectral mixture (SM) kernel \cite{wilson2013gaussian}.
Our goal is to evaluate if \methodname{} produces reliable predictive variances, even with complex kernel functions.

We compute the variances for Exact GP, KISS-GP w/o \methodname{}, and KISS-GP with \methodname{} models with a $10$-mixture SM kernel.
In \autoref{fig:airline_results}, we see that the KISS-GP/\methodname{} confidence intervals match the Exact GP's intervals extremely well.
The SMAE of \methodname{}'s predicted variances (compared against Exact GP variances) is $1.29 \times 10^{-4}$.
Although not shown in the plot, we confirm the reliability of these predictions by computing the log-likelihood of the test data.
We compare the KISS-GP/\methodname{} model to an Exact GP, a KISS-GP model without \methodname{}, and a sparse variational GP (SGPR) model with $m=1000$ inducing points.\footnote{
 Implemented in GPFlow \cite{matthews2017gpflow}} \cite{titsias2009variational,hensman2013gaussian}.
All methods achieve nearly identical log-likelihoods, ranging from $-221$ to $-222$.

\paragraph{Large datasets.}
We measure the accuracy of \methodname{} variances on several large-scale regression benchmarks from the UCI repository \cite{asuncion2007uci}.
We compute the variance for all test set data points.
Each of the models use deep RBF kernels (DKL) on these datasets with the architectures described in \cite{wilson2016deep}.
Deep RBF kernels are extremely flexible (with up to $10^5$ hyperparameters) and are well suited to model many types of functions.
In \autoref{tab:large_dataset_results}, we report the SMAE of the KISS-GP/\methodname{} variances compared against the two baselines.
On all datasets, we find that \methodname{} matches KISS-GP w/o \methodname{} variances to at least $5$ decimal points.
Furthermore, KISS-GP/\methodname{} is able to approximate Exact variances with no more than than $0.1\%$ average error.
For any given test point, the maximum variance error is similarly small (e.g. $\leq \! 2.6\%$ on Skillcraft and $\leq \! 2.0\%$ on PoleTele).
Therefore, using \methodname{} to compute variances results in \emph{almost no loss in accuracy}.

\paragraph{Speedup.}
In \autoref{tab:large_dataset_results} we compare the variance computation speed of a KISS-GP model with and without \methodname{} on the UCI datasets.
In addition, we compare against the speed of SGPR with a standard RBF kernel, a competitive scalable GP approach.
On all datasets, we measure the time to compute variances {\bf from scratch}, which includes the cost of pre-computation (though, as stated in \autoref{sec:method}, this typically occurs during training).
In addition, we report the speed {\bf after pre-computing} any terms that aren't specific to test points (which corresponds to the test time speed).
We see in \autoref{tab:large_dataset_results} that KISS-GP with \methodname{} yields a substantial speedup over KISS-GP without \methodname{}.
The speedup is between $4\times$ and $44\times$, even when accounting for \methodname{}'s precomputation.
At test time after pre-computation, \methodname{} is \emph{up to $2,\!000\times$ faster}.
Additionally, KISS-GP/\methodname{} is significantly faster than SGPR models.
For SGPR models with $m=100$ inducing points, the KISS-GP/\methodname{} model (with $m=10,\!000$ inducing points) is up to $10\times$ faster before pre-computation and $100\times$ faster after.
With $m=1000$ SGPR models, KISS-GP/\methodname{} is up to $20\times$/$500\times$ faster before/after precomputation.
The biggest improvements are obtained on the largest datasets since \methodname{}, unlike other methods, is independent of dataset size at test time.

\begin{figure}[t!]
  \centering
  \includegraphics[width=0.95\columnwidth]{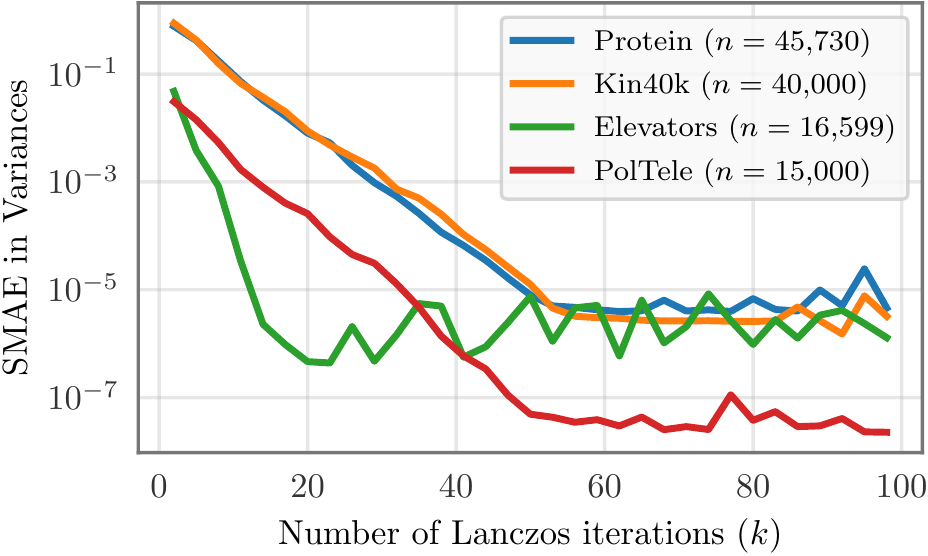}
  \vspace{-2ex}
  \caption{
    Predictive variance error as a function of the Lanczos iterations (KISS-GP model, $m=10,\!000$, Protein, Kin40k, PoleTele, Elevators  UCI datasets).
    \label{fig:lanczos_accuracy}
  }
  \vspace{-1ex}
\end{figure}

\paragraph{Accuracy vs. Lanczos iterations.}
In \autoref{fig:lanczos_accuracy}, we measure the accuracy of \methodname{} as a function of the number of Lanczos iterations ($k$).
We train a KISS-GP model with a deep RBF kernel on the four largest datasets from \autoref{tab:large_dataset_results}, using the setup described above.
We measure the SMAE of KISS-GP/\methodname's predictive variances compared against the standard KISS-GP variances (KISS-GP w/o \methodname).
As seen in \autoref{fig:lanczos_accuracy}, error decreases \emph{exponentially} with the number of Lanczos iterations, up until roughly $50$ iterations.
After roughly $50$ iterations, the error levels off, though this may be an artifact of floating-point precision (which may also cause small subsequent fluctuations).
Recall that $k$ also corresponds with the rank of the $R$ and $R'$ matrices in \eqref{eq:pred_covar_ski_fast}.

\subsection{Sampling}

We evaluate the quality of posterior samples drawn with KISS-GP/\methodname{} as described in \autoref{sec:sampling_method}.
We compare these samples to three baselines: sampling from an {\bf Exact GP} using the Cholesky decomposition, sampling from an {\bf SGPR} model with Cholesky, and sampling with random {\bf Fourier features} \citep{rahimi2008random} -- a method commonly used in Bayesian optimization \cite{hernandez2014predictive,wang2017max}.
The KISS-GP/\methodname{} and SGPR models use the same architecture as described in the previous section.
For Fourier features, we use 5000 random features -- the maximum number of features that could be used without exhausting available GPU memory.
We learn hyperparameters on the Exact GP and then use the same hyperparameters for the all methods.

We test on the two largest UCI datasets which can still be solved exactly (PolTele, Eleveators) and two Bayesian optimization benchmark functions (Eggholder -- 2 dimensional, and Styblinski-Tang -- 10 dimensional).
The UCI datasets use the same training procedure as in the previous section.
For the BayesOpt functions, we evaluate the model after 100 queries of max value entropy search \cite{wang2017max}.
We use a standard (non-deep) RBF kernel for Eggholder, and for Syblinski-Tang we use the additive kernel decomposition suggested by~\citet{kandasamy2015high}.\footnote{
  The Syblinski-Tang KISS-GP model uses the sum of 10 RBF kernels -- one for each dimension -- and $m=100$ inducing points.
}

\paragraph{Sample accuracy.}
In \autoref{tab:sampling_results} we evaluate the accuracy of the different sampling methods.
We draw $s\!=\!1000$ samples at $t\!=\!10,\!000$ test locations and compare the sample covariance matrix with the true posterior covariance matrix $k_{f\mid\dset}(\X^{*},\X^{*'})$ in terms of element-wise mean absolute error.
It is worth noting that all methods incur some error -- even when sampling with an Exact GP.
Nevertheless, Exact GPs, SGPR, and \methodname{} produce very accurate sample covariance matrices.
In particular, Exact GPs and \methodname{} achieve between 1 and 3 orders of magnitude less error than the random Fourier Feature method.

\paragraph{Speed.}
Though \methodname{}, Exact GPs, and SGPR have similar sample accuracies, \methodname{} is much faster.
Even when accounting for \methodname's pre-computation time (i.e. sampling ``from scratch''), \methodname{} is comparable to Fourier features and SGPR in terms of speed.
At test time (i.e. after pre-computation), \methodname{} is up to $18,\!000$ times faster than Exact.

\paragraph{Bayesian optimization.}
Many acquisition functions in Bayesian optimization rely on sampling from the posterior GP.
For example, max-value entropy search \cite{wang2017max} draws samples from a posterior GP in order to estimate the function's maximum value $p(y^{*} \! \mid \! \dset)$.
The corresponding maximum \emph{location} distribution, $p(x^{*} \! \mid \! \dset)$, is also the primary distribution of interest in predictive entropy search \cite{hernandez2014predictive}.
\begin{figure}[t!]
  \centering
  \includegraphics[width=\columnwidth]{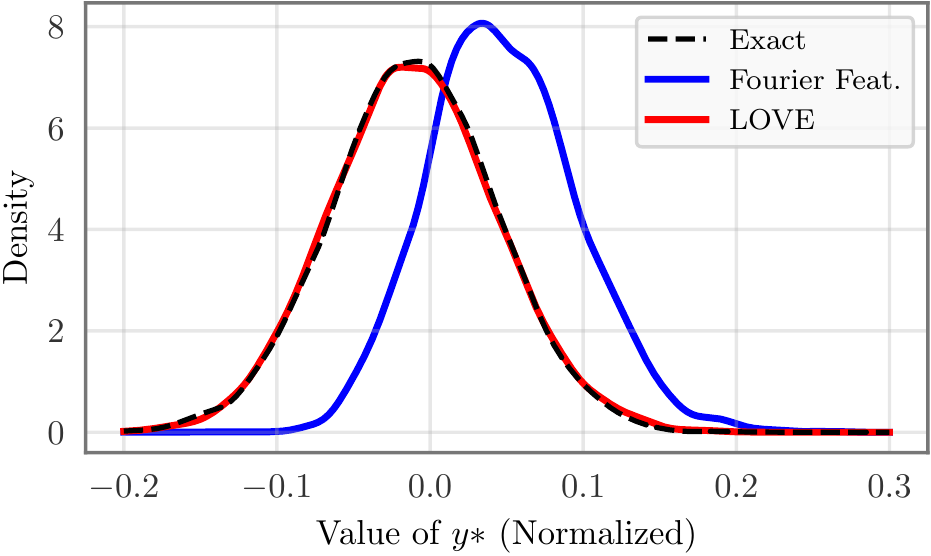}
  \vspace{-4ex}
  \caption{
    A density estimation plot of the predictive maximum distribution, $p(y^* \! \mid \! \dset)$ for the (normalized) Eggholder function after 100 iterations of BayesOpt.
    Samples drawn with KISS-GP/\methodname{} closely match samples drawn using an Exact GP. See \citet{wang2017max} for details.
    \label{fig:thompson_sampling_comparison}
  }
    \vspace{-2ex}
\end{figure}
In \autoref{fig:thompson_sampling_comparison}, we visualize each method's sampled distribution of $p(y^{*} \! \mid \! \dset)$ for the Eggholder benchmark.
We plot kernel density estimates of the sampled maximum value distributions after $100$ BayesOpt iterations.
Since the Exact GP sampling method uses the exact Cholesky decomposition, its sampled max-value distribution can be considered closest to ground truth.
The Fourier feature sampled distribution differs significantly.
In contrast, \methodname{}'s sampled distribution very closely resembles the Exact GP distribution, yet
\methodname{} is $700 \times$ faster than the Exact GP on this dataset (\autoref{tab:sampling_results}).

  %!TEX root=../main.tex
\section{Discussion, Related Work, and Conclusion}
\label{sec:discussion}

This paper has primarily focused on \methodname{} in the context of KISS-GP as its underlying inducing point method.
This is because of KISS-GP's accuracy, its efficient MVMs, and its constant asymptotic complexities when used with \methodname{}.
However, LOVE and MVM inference are fully compatible with other inducing point techniques as well. Many inducing point methods make use of the subset of regressors (SOR) kernel approximation $\K_{XX} \approx
\K_{XU}\K_{UU}^{-1}\K_{UX}$, optionally with a diagonal correction \cite{snelson2006sparse}, and focus on the problem of learning the
inducing point locations \cite{quinonero2005unifying,titsias2009variational}. After $\bigo{m^{3}}$ work to Cholesky decompose
$\bigo{\K_{UU}}$, this approximate kernel affords $\bigo{n + m^{2}}$ MVMs. One could apply LOVE to these methods and
compute a test-invariant cache in $\bigo{knm+km^{2}}$ time, and then compute single predictive covariances in $\bigo{mk}$ time.
We note that, since these inducing point methods make a rank $m$ approximation to the kernel, setting $k\!=\!m$ produces exact solves with Lanczos decomposition, and recovers the $\bigo{nm^{2}}$ precomputation time and $\bigo{m^2}$ prediction time of these methods.

\paragraph{Ensuring Lanczos solves are accurate.}
Given a matrix $\widehat K_{XX}$, the Lanczos decomposition $Q_k T_k Q_k^\top$ is designed to approximate the solve $\widehat K_{XX}^{-1} \bb$, where $\bb$ is the first column of $Q_k$.
As argued in \autoref{subsec:lanczos}, the $Q_k$ and $T_k$ can usually be re-used to approximate the solves $\widehat K_{XX}^{-1} (W_X^\top K_{UU}) \approx Q_k T_k^{-1} Q_k^\top (W_X^\top K_{UU})$.
This property of the Lanczos decomposition is why \methodname{} can compute fast predictive variances.
While this method usually produces accurate solves, the solves will not be accurate if some columns of $(W^\top_X K_{UU})$ are (nearly) orthogonal to the columns of $Q_k$.
In this scenario, \citet{saad1987lanczos} suggests that the additional Lanczos iterations with a new probe vector will correct these errors.
In practice, we find that these countermeasures are almost never necessary with \methodname{} -- the Lanczos solves are almost always accurate.

\paragraph{Numerical stability of Lanczos.}
A practical concern for \methodname{} is round-off errors that may affect the Lanczos algorithm.
In particular it is common in floating point arithmetic for the vectors of $Q$ to lose orthogonality \cite{paige1970practical,simon1984lanczos,golub2012matrix}, resulting in an incorrect decomposition.
To correct for this, several methods such as full reorthogonalization and partial or selective reorthogonalization exist \cite{golub2012matrix}.
In our implementation, we use full reorthogonalization when a loss of orthogonality is detected.
In practice, the cost of this correction is absorbed by the parallel performance of the GPU and does not impact the final running time.

\paragraph{Conclusion.}
%We have demonstrated a method for computing predictive covariances and drawing samples from the predictive distribution in constant time with almost no loss in accuracy.
Whereas the running times of previous state-of-the-art methods depend on dataset size, \methodname{} provides \emph{constant time} and near-exact predictive variances.
In addition to providing scalable predictions, \methodname{}'s fast sampling procedure has the potential to dramatically simplify a variety of GP applications such as \citep[e.g.,][]{deisenroth2011pilco,hernandez2014predictive,wang2017max}.
These applications require fast posterior samples, and have previously relied on parametric approximations or finite basis approaches for approximate sampling \citep[e.g.,][]{deisenroth2011pilco,wang2017max}.
The ability for \methodname{} to obtain samples in linear time and variances in constant time will improve these applications, and may even unlock entirely new applications for Gaussian processes in the future.

  \section*{Acknowledgements}
  JRG, GP, and KQW are supported in part by the III-1618134, III-1526012,
  IIS-1149882, IIS-1724282, and TRIPODS-1740822 grants from the National Science
  Foundation. 
  In addition, they are supported by the Bill and Melinda Gates Foundation, the Office of Naval Research, and SAP America Inc.
  AGW and JRG are supported by NSF IIS-1563887.

  \bibliography{citations}
  \bibliographystyle{icml2018}
\end{document}